\documentclass{article}

     \usepackage[preprint]{neurips_2020}

\usepackage[utf8]{inputenc} 
\usepackage[T1]{fontenc}    
\usepackage{hyperref}       
\usepackage{url}            
\usepackage{booktabs}       
\usepackage{amsfonts}       
\usepackage{nicefrac}       
\usepackage{microtype}      
\usepackage{natbib}
\usepackage{sidecap}
\usepackage{float}
\usepackage{stmaryrd}
\usepackage{graphicx}
\usepackage{floatflt}
\usepackage{relsize}
\usepackage{color, colortbl}
\usepackage{multirow}
\usepackage{amsmath}
\usepackage{wrapfig}
\usepackage{arydshln}
\usepackage{enumitem}

\usepackage{bbm}
\usepackage{adjustbox}
\usepackage{caption}
\usepackage{subcaption}
\usepackage{verbatimbox}
\usepackage{multirow}
\usepackage{color, colortbl}
\usepackage{stfloats}

\title{Improving Graph Neural Network Representations \\ of Logical Formulae with Subgraph Pooling}

\author{%
  Maxwell Crouse$^{*\dagger}$ \ \ \ \ Ibrahim Abdelaziz$^\ddag$ \ \ \ \ Cristina Cornelio$^\ddag$ \ \ \ \ \\ {\bf Veronika Thost$^\S$ \ \ \ \ Lingfei Wu$\ddag$ \ \ \ \ Kenneth Forbus$\dagger$ \ \ \ \ Achille Fokoue$^\ddag$} \\
  $^\dagger$Qualitative Reasoning Group, Northwestern University \\
  $^\ddag$IBM T.J. Watson Research Center, IBM Research \\
  $^\S$MIT-IBM Watson AI Lab, IBM Research
}

\begin{document}

\maketitle

\begin{abstract}
Recent advances in the integration of deep learning with automated theorem proving have centered around the representation of logical formulae as inputs to deep learning systems. In particular, there has been a growing interest in adapting structure-aware neural methods to work with the underlying graph representations of logical expressions. While more effective than character and token-level approaches, graph-based methods have often made representational trade-offs that limited their ability to capture key structural properties of their inputs. In this work we propose a novel approach for embedding logical formulae that is designed to overcome the representational limitations of prior approaches. Our architecture works for logics of different expressivity; e.g., first-order and higher-order logic. 
We evaluate our approach on two standard datasets and show that the proposed architecture achieves state-of-the-art performance on both premise selection and proof step classification.

\end{abstract}

\section{Introduction}
\label{sec:intro}

Automated theorem proving studies the design of automated systems that reason over 
logical theories (collections of \emph{axioms} that are formulae known to be true) to generate formal proofs of given conjectures. It has been a longstanding, active area of artificial intelligence research that has demonstrated utility in the design of operating systems \cite{klein2009operating,klein2014comprehensive}, distributed systems \cite{garland1998ioa,hawblitzel2015ironfleet}, compilers \cite{curzon1991verified,leroy2009formal}, microprocessor design \cite{hunt1989microprocessor}, and in general mathematics \cite{hales2017formal}.

Classical automated theorem provers (ATPs) have historically been most useful for solving problems that require complex chains of reasoning steps to be executed over smaller sets of axioms (see 
TPTP \cite{sutcliffe2009tptp} for examples). When faced with problems for which thousands to millions of axioms are provided (only a handful of which may be needed at a time), even state-of-the-art theorem provers have 
difficulty \cite{ramachandran2005first,hoder2011sine}. This deficiency has become more evident in recent years, as large logical theories
\cite{matuszek2006introduction,mizar40for40,pease2002suggested} have become more widely available. A natural way to scale ATPs to broader domains has been to design sophisticated mechanisms that allow them to determine which axioms or intermediate proof outputs merit exploration in the proof search process. These mechanisms thus prune an otherwise unmanageably large proof search space down to a size that can be handled efficiently by classical theorem provers. 
The task of classifying axioms as being useful 
to prove a given conjecture is referred to as \emph{premise selection}, while the task of classifying intermediate proof steps as being a part of a successful proof for a conjecture is referred to as \emph{proof step classification}. 

Initial approaches for solving these two tasks proposed heuristics based on simple symbol co-occurrences between formulae~\cite{hoder2011sine,roederer2009divvy,kuhlwein2012overview}. While effective, these methods were soon surpassed by machine-learning techniques which could automatically adjust themselves to the needs of particular domains~\cite{alama2014premise,irving2016deepmath}. At present, there has been a rising interest in developing 
neural approaches for both premise selection and proof step classification \cite{bansal2019holist,kaliszyk2017holstep,loos2017deep}; however, the complex and structured nature of logical formulae has made this development 
challenging.
Neural approaches that take into account a formula's structure (e.g., its parse tree), have been shown to outperform their more basic counterparts which operate on only a formula's symbols \cite{wang2017premise,paliwal2019graph}. The two most commonly used structure-aware neural methods have been Tree LSTMs \cite{tai2015improved} and GNNs \cite{kipf2016semi}. However, as they have been applied in this domain, these methods appear to be leaving out useful structural information.

When used to embed the parse tree of a logical formula, Tree LSTMs generate embeddings that represent the parse tree globally, but they miss logically important information like shared subexpressions and variable quantifications.
Conversely, traditional GNN approaches appear to better capture shared subexpressions and variable quantifications; however, the global graph embedding they produce for the whole formula consists of a simple pooling operation over individual node embeddings; each representing only themselves and their local neighborhoods. 
Additionally, most prior approaches have embedded the premise and conjecture formulae independently of each other \cite{wang2017premise,loos2017deep,paliwal2019graph,irving2016deepmath}. They first embed the graph of the premise and then separately embed the conjecture graph, resulting in the contents of one formula having no influence on the embedding of the other.
%
%

To address these issues, we present a novel, two-phase embedding approach that operates over the DAG representations of logical formulae and is designed with careful consideration to their particular properties. Our method first produces an initial set of high-quality embeddings for nodes that incorporates more than just their local neighborhoods. Then, it pools the embeddings together in a structure-dependent way to generate a single graph-level embedding. 
This decoupling provides a clear point at which information between formulae can be exchanged, which allows us to define a localized attention-based exchange mechanism that can regulate information flow between the concurrent formula embedding processes.

We demonstrate the effectiveness and generality of our approach by evaluating classification performance on two standard datasets that involve different logical formalisms; the Mizar dataset \cite{mizar40for40,irving2016deepmath} for first-order logic and the Holstep dataset \cite{kaliszyk2017holstep} for higher-order logic. Our experiments show that the architecture introduced in this paper outperforms all previous approaches on the binary classification tasks of premise selection and proof step classification for both datasets. We also demonstrate how to easily integrate our approach with \emph{E}, a well-established theorem prover \cite{schulz2013system}, as its premise selection mechanism, allowing it to find more proofs (61.6\% improvement) in a large-theory setting.

To summarize, our main contributions are:
{\bf 1)} We show how to leverage the DAG structure implicit in logical formulae to produce more effective embeddings than traditional approaches operating over the local neighborhoods of individual nodes; 
{\bf 2)} We introduce a novel neural architecture that employs a localized attention mechanism to allow formulae to exchange information during the embedding process;
{\bf 3)} We provide an extensive series of experiments and compare a range of neural architectures, showing significant improvement over existing state-of-the-art methods on two standard ATP classification datasets.

\section{Related Work}\label{sec:related_work}


We note that premise selection and proof step classification are not intrinsically machine learning tasks. The earliest approaches to premise selection \cite{hoder2011sine} were simple heuristics capturing the (transitive) co-occurrence of symbols in a given axiom and conjecture. Soon after, it was recognized that machine-learning techniques would be effective tools for solving this problem. The work of \cite{alama2014premise} introduced a kernel method for premise selection where the similarity between two formulae was computed by the number of common subterms and symbols. Deepmath \cite{irving2016deepmath} was the first deep learning approach 
to this problem, comparing the performance of sequence models over character and symbol-level representations of logical formulae. In \cite{kucik2018premise}, the authors proposed a symbol-level method that learned low-dimensional distributed representations of function symbols and used those to construct embedded representations of given formulae for premise selection. The work of \cite{olvsak2019property} introduced a GNN for representing specifically first-order logic formulae in conjuctive normal form that captured certain logical invariances like reorderings of clauses and literals. 

Recently, Holstep \cite{kaliszyk2017holstep}, a new formal dataset designed to be large enough to evaluate neural methods for premise selection and proof step classification (among other tasks), was introduced.
Along with the dataset came a set of benchmark deep learning models that operated over character and symbol-level representations of higher-order logic formulae. FormulaNet \cite{wang2017premise} was the first approach to transform a formula into a rooted DAG 
(a modified version of the parse tree) and then process the resulting graph 
with a GNN. Their GNN produced embeddings for each node within a formula's graph representation and then max pooled across node embeddings to get a formula-level embedding.

There are several other related works in this area that focus on different tasks (e.g., proof guidance, the combined, online version of the aforementioned two tasks). 
Deep learning approaches to proof guidance include \cite{loos2017deep}, where the authors explored a number of neural architectures in their implementation (including a Tree LSTM that encoded parse trees of logical formulae). \cite{paliwal2019graph} represented formulae as DAGs with shared subexpressions and used message-passing GNNs (MPNNs) to generate embeddings that could be used to guide theorem proving on the higher-order logic benchmark of \cite{bansal2019holist}. However, like \cite{wang2017premise}, the graph-level embeddings produced by their approach were simple, consisting only of a max pooling over individual node embeddings. The work of \cite{evans2018can} introduced a dataset for evaluating neural models on entailment for propositional logic and explored the use of several popular neural architectures on the proposed task. 
\cite{huang2018gamepad} where the authors introduced the GamePad dataset for evaluating neural models on the tasks of position evaluation and tactic prediction.

\begin{figure*}[t]
\centering
    \footnotesize
    \includegraphics[width=0.85\textwidth]{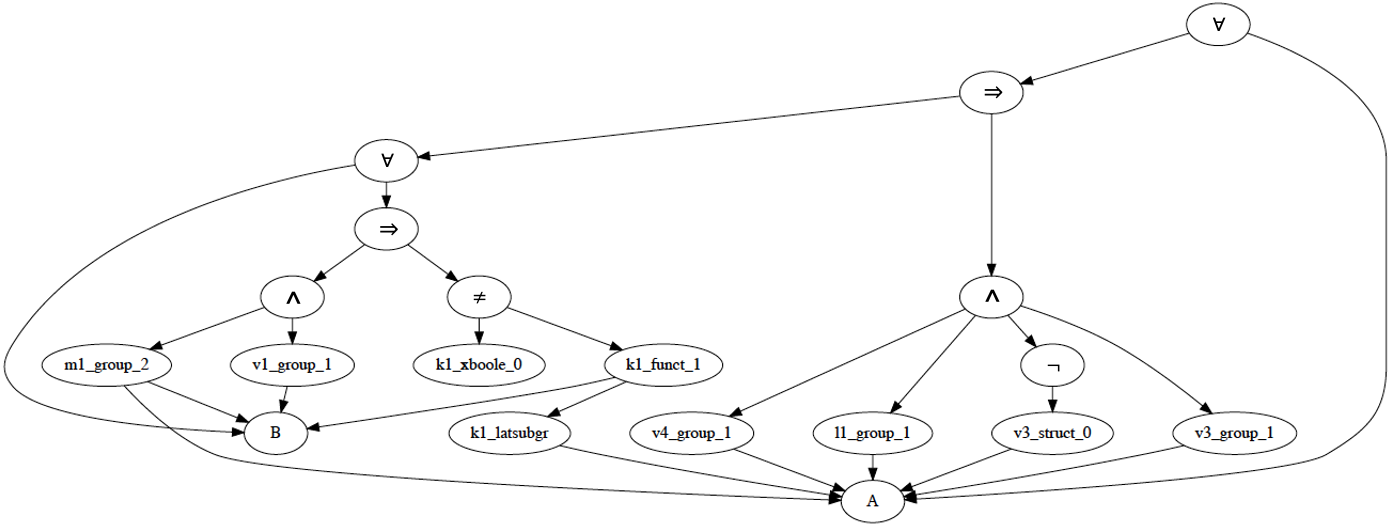}
    \caption{Graph representation for a conjecture which regards the lattice of subgroups of a group}
    \label{fig:formula}
    \vspace{-0.1in}
\end{figure*}

\section{Formula Representation}
\label{sec:formulas}

\paragraph{Background:} {\em First-order logic formulae} are formal expressions based on an alphabet of predicate, function, and variable symbols which are combined by logical connectives.
A term is either a 
variable, a constant (function with no arguments), or, inductively, a function applied to a tuple of terms. A formula is either a predicate applied to a tuple of terms or, inductively, a connective (e.g., $\wedge$ read as ``and'') 
applied to some number of formulae. 
In addition, variables in formulae can be quantified by quantifiers (e.g., by $\forall$ read as ``for all''), where a quantifier introduces an additional semantic restriction for the interpretation of the variables it quantifies.
{\em Higher-order logic formulae} also allow for quantification over predicate and function symbols or the application of predicates over other predicates. For more details on both first and higher-order logic, we refer the reader to \cite{taylor1999practical}.


\paragraph{Logical Formulae as Graphs:} While the earliest work on integrating deep learning with reasoning techniques used symbol- or word-level representations of input formulae \cite{irving2016deepmath,kaliszyk2017holstep} (considering formula strings as words), subsequent work explored using formula parse trees \cite{loos2017deep,evans2018can,huang2018gamepad} or rooted DAG forms \cite{wang2017premise,paliwal2019graph}. On the Holstep \cite{kaliszyk2017holstep} and Holist \cite{bansal2019holist} datasets, the DAG forms of logical formulae were found to be the more useful than bag-of-symbols and tree-structured encodings \cite{wang2017premise,paliwal2019graph}. We focus on rooted-DAG representations of formulae; Figure~\ref{fig:formula} shows an unmodified example of such a representation. The DAG associated to a formula corresponds to its parse tree, where directed edges are added from parents to their arguments and shared subexpressions are mapped to the same subgraphs. As in \cite{wang2017premise}, all instances of the same variable are collapsed into a single node (which maintains all prior connections) and the name of each variable is replaced by a generic variable token.

\paragraph{Edge Labeling:}
\label{sec:edge_labels}
Capturing the ordering of arguments of logical expressions is still an open topic of research. \cite{wang2017premise} used a so-called \emph{treelet} encoding scheme that represents 
the position of a node relative to other arguments of the same parent as triples. \cite{paliwal2019graph} used positional edge labels, assigning to each edge a label reflecting the position of its target node in the argument list of the node's parent.
We follow the latter strategy, albeit, with modifications. In our formulation, edge labels are determined by a \textit{partial ordering}. For unordered logical connectives (e.g., $\wedge$, $\Leftrightarrow$) and predicates (e.g., $=$) all arguments are of the same rank. For other predicates, functions, and logical connectives 
the arguments are instead linearly ordered. However, we also support hybrid cases like simultaneous quantification over multiple variables. 
The label given to each argument edge in the graph is the rank of the corresponding argument with respect to the parent concatenated with the type of the parent.


%

\section{Our Approach}
\label{sec:reachability}
In this work, we broadly distinguish between node embedding methods by \textit{reachability}. More formally, consider a binary adjacency relation $\mathcal{R}$ defined for a set of graphs $\mathcal{G}$. The $k$-reachability relation $\mathcal{R}^k$ is given as the $k$-th power of $\mathcal{R}$, which is defined recursively with $\mathcal{R}^k = \mathcal{R}^{k - 1} \circ \mathcal{R}$ and $\mathcal{R}^1 = \mathcal{R}$. We can define the transitive closure of $\mathcal{R}$ as simply $\mathcal{R}^+ = \mathcal{R}^\infty$. Letting the set of all nodes be $\bar{V} = \bigcup_{G = (V, E) \in \mathcal{G}} V$, we say that a graph embedding function $f$ is a $\mathcal{R}^k$ embedding method if there exists a $k \in \mathbb{N}$ such that $\mathcal{R}^k \neq \mathcal{R}^+$ where for every $u$ in $\bar{V}$ we have that $f$ computes the value of $u$ as a function of \textit{only} the embeddings for $\{ v \in \bar{V} \ | \ \mathcal{R}^k(u, v) \}$. Naturally, we define a $\mathcal{R}^+$ embedding method as one for which the opposite holds, i.e. for each $u$ we have that $f$ computes the value of $u$ as a function of the embeddings for all $v$ where $\mathcal{R}^+(u, v)$ holds. This distinction is particularly useful to make for graphs in the logic domain, as the transitive closure of adjacency is necessary for many key logical operations. As a trivial but important example, consider checking for the resolvability or unifiability of two ground formulas. Potentially all nodes of the two formulas would need to be examined, meaning that if both formulas had depth $> k$ then a procedure defined with $\mathcal{R}^k$ that checks only some subset of nodes within a fixed range of the root would be insufficient.

We view $\mathcal{R}^+$ embedding methods as those that perform a sophisticated type of subgraph pooling. That is, a $\mathcal{R}^+$ node embedding method computes the embedding for a node $u$ as a function of the embeddings of all nodes reachable from $u$, i.e. a pooling of all such node embeddings. By definition, they incorporate as much graph context as is possible (i.e., the transitive closure of $\mathcal{R}$). While $\mathcal{R}^+$ node embedding methods naturally lend themselves to graph-level readout functions (and we will also use them in this way), we note that these concepts are defined for node-level embeddings (an important distinction to make, as for certain applications the input graphs could be disconnected).

\begin{figure*}
\centering
\includegraphics[width=0.6\textwidth]{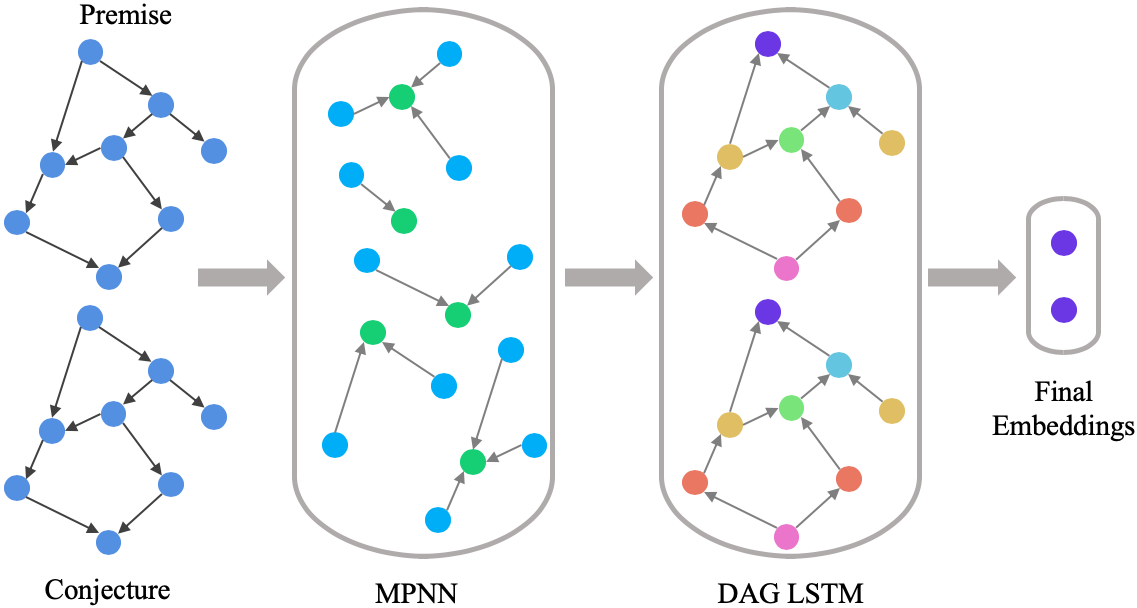}
\caption{A depiction of the overall embedding process with an MPNN as the initial node embedder and DAG LSTM as the pooling method. In both boxes, arrows indicate flow of information. 
}
\label{fig:architecture}
\vspace{-0.1in}
\end{figure*}

Our approach operates in two stages (see Figure \ref{fig:architecture}). First, a neural network generates embeddings for each node of an input formula's graph representation. Then, the node embeddings are passed into a follow-up $\mathcal{R}^+$ embedding method, referred to as the pooling method, that has $\mathcal{R}$ as the parent relation. The embedding for the root node of the input formula is returned, which is a function of all nodes in its graph. Our approach is very modular, with any node-level embedding method capable of serving as the initial node embedder (though we are mainly interested in $\mathcal{R}^+$ embedding methods) and any $\mathcal{R}^+$ embedding method being usable as the pooling method. Thus, in the next sections we describe the node embedding methods independently, and then we describe the classification process.

\subsection{$\mathcal{R}^{k}$ Embedding Methods}
\label{sec:local_embedders}
\paragraph{Message-Passing Graph Neural Networks:} The MPNN framework can be thought of as an iterative update procedure that represents a node as an aggregation of information from its local neighborhood. 
To begin, our MPNN assigns each node $v$ and edge $e$ of the input graph $G = (V, E)$ an initial embedding, $x_v$ and $x_e$. Then, following \cite{wang2017premise}, initial node states are computed by passing each such embedding through batch normalization \cite{ioffe2015batch} and a ReLU, producing node states $h_v^{(0)} = F_{V}(x_v)$ and edge states $h_e = F_{E}(x_e)$. Lastly, a message-passing phase runs for $t = 1, \ldots, k$ rounds
\begin{equation*}
\begin{gathered}
m_{v_p}^{(t)} = \sum_{w \in \mathcal{P}_{\mathcal{A}}(v)} F^{(t)}_{M_A}\big([h_v^{(t - 1)} || h_w^{(t - 1)} || h_{e_{vw}}]\big), \ \ \ \ \ 
m_{v_c}^{(t)} = \sum_{w \in \mathcal{P}_{\mathcal{C}}(v)} F^{(t)}_{M_C}\big([h_v^{(t - 1)} || h_w^{(t - 1)} || h_{e_{vw}}]\big) \\
h_v^{(t)} = h^{(t - 1)}_v + F^{(t)}_{A}\big([h_v^{(t - 1)} || m_{v_p}^{(t)} || m_{v_c}^{(t)} ]\big)
\end{gathered}
\end{equation*}
where $\mathcal{P}_{\mathcal{A}}$ and $\mathcal{P}_{\mathcal{C}}$ are functions that take a node $v$ and return the immediate ancestors / children of $v$ in $G$, and $F^{(t)}_{M_A}$, $F^{(t)}_{M_C}$, and $F^{(t)}_{A}$ are feed-forward networks unique to the $t$-th round of updates, and $||$ denotes vector concatenation. 
The reachability relation $\mathcal{R}$ in this context is defined as $\mathcal{R}(u, v) = \mathcal{A}(u, v) \vee \mathcal{C}(u, v)$ where $\mathcal{A}$ and $\mathcal{C}$ are relations that hold true for immediate ancestor and child relationships, respectively.
Similar to \cite{gilmer2017neural}, $m_{v_p}^{(t)}$ and $m_{v_c}^{(t)}$ should be considered the \textit{messages} to be passed to $h_v$, and $h_v^{(t)}$ represents the node embedding for node $v$ after $t$ rounds of iteration.

\paragraph{Graph Convolutional Neural Networks: } Like with our MPNNs, for our Graph Convolutional Networks (GCNs) \cite{kipf2016semi}, the reachability relation $\mathcal{R}$ is given as the undirected adjacency relation, i.e., for nodes $u$ and $v$ we have $\mathcal{R}(u, v) = \mathcal{A}(u, v) \vee \mathcal{C}(u, v)$. First, each node $v \in V$ is associated with an embedding $h_v$. Then, for $t = 1, \ldots, k$ rounds, updated embeddings are computed as
\begin{alignat*}{2}
h_{v}^{(t)} &= \phi \big( W^{(t)} \big( \dfrac{h_v^{(t - 1)}}{|\mathcal{N}(v)|} + \sum_{w \in \mathcal{N}(v)} \dfrac{h_{w}^{(t - 1)}}{\sqrt{|\mathcal{N}(v)||\mathcal{N}(w)|}} \big) \big)
\end{alignat*}
where $\phi$ is a non-linearity (in this work, we use a ReLU), $\mathcal{N}(u) = \mathcal{P}_{\mathcal{A}}(u) \cup \mathcal{P}_{\mathcal{C}}(u)$, and $W^{(t)}$ is a learned matrix specific to the $t$-th round of updates.

\subsection{$\mathcal{R}^{+}$ Embedding Methods}
\label{sec:global_embedders}
\paragraph{DAG LSTMs:} DAG LSTMs can be viewed as the generalization of Tree LSTMs \cite{tai2015improved} to DAG-structured data. With initial node embeddings $s_v$, the DAG LSTM uses the same N-ary equations as the Tree LSTM to compute node states $h_v$
\begin{alignat*}{4}
    &i_v &&= \sigma \big( W_i s_v + \sum_{w \in \mathcal{P}_\mathcal{R}(v)} U_i^{(e_{vw})} h_w + b_i \big) \ \ \ \ \ \ \ &&f_{vw} &&= \sigma \big( W_f s_v + U_f^{(e_{vw})} h_w + b_f \big) \\
    &o_v &&= \sigma \big( W_o s_v + \sum_{w \in \mathcal{P}_\mathcal{R}(v)} U_o^{(e_{vw})} h_w + b_o \big) \ \ \ \ \ \ \ &&c_{v} &&= i_v \odot \hat{c}_v + \sum_{w \in \mathcal{P}_\mathcal{R}(v)} f_{vw} \odot c_{w} \\
    &\hat{c}_v &&= \tanh{ \big( W_c s_v + \sum_{w \in \mathcal{P}_\mathcal{R}(v)} U_c^{(e_{vw})} h_w + b_c \big) } \ \ \ \ \ \ \ &&h_{v} &&= o_v \odot \tanh{\big(c_v\big)}
\end{alignat*}
where $\odot$ denotes element-wise multiplication, $\sigma$ is the sigmoid function and $U_i^{(e_{vw})}$, $U_o^{(e_{vw})}$, $U_c^{(e_{vw})}$, and $U_f^{(e_{vw})}$ are learned matrices (different for each edge type). $i$ and $o$ represent input and output gates, while $c$ and $\hat{c}$ are intermediate computations (memory cells), and $f$ is a forget gate that modulates the flow of information from individual arguments into a node's computed state. $\mathcal{P}_\mathcal{R}$ is a predecessor function that returns the set of nodes for which $\mathcal{R}$ holds true, i.e. $\mathcal{P}_\mathcal{R}(u) = \{ v \in V \ | \ \mathcal{R}(u, v) \}$. In this work, it returns either the parents or the children, depending on whether the direction of accumulation is desired to go upwards or downwards. For readability, we omitted the layer normalization \cite{ba2016layer} applied to each matrix multiplication (e.g., $W_i s_v$, $U_i h_w$, etc.) from the above equations. Each instance of layer normalization maintained its own separate parameters.

The DAG LSTM we propose here is nearly the same as the Tree LSTM of \cite{tai2015improved}, however there are key implementational differences between the two approaches. In Tree LSTMs, $\mathcal{P}_\mathcal{R}$ typically returns child nodes (since a node can have only one parent), while in our work it can return either children or parents. In addition, batching together node updates in a Tree LSTM can be done at the level of depth (i.e., all nodes at the same depth in the tree can have their updates computed simultaneously); however, with DAGs this batching strategy could cause a node to be updated and overwritten multiple times. To solve this, we propose the use of \emph{topological batching}. In our approach, node updates are computed in the order given by a topological sort of the graph, starting from the leaves (or root depending on $\mathcal{P}$), with updates batched together at the level of topological equivalence, i.e., every node with the same rank can have the updates computed simultaneously.

\paragraph{Attention-Enhanced DAG LSTMs:} In order to allow the contents of the premise and conjecture to influence one another during the embedding process, we introduce a localized attention mechanism that exchanges information between the two graph embeddings. 
Let $S_P$ and $S_C$ be the sets of node embeddings computed by some initial node embedder for the premise and conjecture graphs. Let $\mathcal{I}$ be a function that takes a node and either $S_P$ or $S_C$ and returns all node embeddings from the set where the associated node has an identical label to the given node, i.e. $\mathcal{I}(u, S_C) = \{ s_v \in S_C | u \equiv v\}$. Our approach computes multi-headed attention scores \cite{vaswani2017attention} between identically labeled nodes and uses those attention scores to build new embeddings that provide cross graph information to the pooling procedure. For an input $u$, for each $k_j \in \mathcal{I}(u, S_C)$ we compute
\begin{gather*}
    \hat{q}_{i} = W^{(q)}_{i} s_u,\;\;\;\;k_{ij} = W^{(k)}_{i} k_{j},\;\;\;\;v_{ij} = W^{(v)}_{i} k_j
\end{gather*}
where $W^{(q)}_{i}$, $W^{(k)}_{i}$, and $W^{(v)}_{i}$ are learned matrices for each of the $i = 1, \ldots, N$ attention heads.
\begin{gather*}
    w_{ij} = \dfrac{\hat{q}_{i}^\top k_{ij}}{\sqrt{b_{\hat{q}}}},\;\;\;\;\alpha_{ij} = \dfrac{\exp{(w_{ij})}}{\sum_{j^\prime}\exp(w_{ij^\prime})}, 
\end{gather*}
where $b_{\hat{q}}$ is the dimensionality of $\hat{q}_i$ and $\alpha_{ij}$ is computed by the standard softmax
\begin{gather*}
    q_{i} = \sum_{j} \alpha_{ij} v_{ij},\;\;\;\;s_u^\prime = \sigma(W^{(g)}r_u) \odot ( W^{(o)} \bigparallel_{i = 1}^{N} q_i)
\end{gather*}
The final vector $s_u^\prime$ for input $s_u$ is a combination of its $N$ transformations, with $W^{(g)}$ and $W^{(o)}$ being learned matrices, $r_u$ a learned vector for the type (e.g., quantifier, predicate, etc.) of node $u$, and $\bigparallel$ denoting vector concatenation over the $N$ attention vectors. The gating mechanism $\sigma\big(W^{(g)} r_u)$ we propose here simply allows for the architecture to cut off information flow between the two graphs if doing so improves loss, thus turning the architecture into the simpler DAG LSTM introduced previously.
Following the attention computation, each $s_u$ is replaced by $\bar{s}_u = s_u || s^\prime_u$ and a DAG LSTM then processes each node embedding.


\paragraph{Bidirectional DAG LSTMs:} We also explore a simple extension of the DAG LSTM from above to a Bidirectional DAG LSTM. This extends the $\mathcal{R}$ relation from being either ancestors or children to being both, i.e. $\mathcal{R}(u, v) = \mathcal{A}(u, v) \vee \mathcal{C}(u, v)$. For a node $u$ and graph $G$, the embedding $s_u$ is
\begin{alignat*}{2}
s_u &= F_{BD}\big( [ \textrm{DAG-LSTM}^{\uparrow}(u, G) \ || \ \textrm{DAG-LSTM}^{\downarrow}(u, G)] \big)
\end{alignat*}
where $F_{BD}$ is a feed-forward network, and DAG-LSTM$^{\uparrow}$ / DAG-LSTM$^{\downarrow}$ are both DAG LSTMs (following the design presented before) which set $\mathcal{R}$ as $\mathcal{A}$ and $\mathcal{C}$, respectively.

\subsection{Classification Process}

In our approach, the final graph embeddings for the premise and conjecture are taken to be the embeddings for the root nodes of the premise and conjecture, $s_P = h^{P}_{root}$ and $s_C = h^{C}_{root}$. For ablation experiments using \textit{only} local neighborhood-based node embedders (MPNN / GCN from Section \ref{sec:local_embedders}), the inputs to the classifier network would be a max pooling of the individual node embeddings for each graph. 
In either case,
the two graph embeddings are concatenated and passed to a classifier feed-forward network $F_{CL}$ for the final prediction $F_{CL}([s_P ; s_C])$.

\section{Experiments and Results}
In this section, we evaluate the performance of our approach to show 1) how accurately can it predict the label of an axiom or proof step, 2) an ablation study that shows the effect of different node embedding and pooling mechanisms, and 3) its impact when integrated with an existing theorem prover in terms of number found proofs. We compare our approach to prior works using two standard datasets: Mizar\footnote{https://github.com/JUrban/deepmath} \cite{mizar40for40} and Holstep\footnote{http://cl-informatik.uibk.ac.at/cek/holstep/} \cite{kaliszyk2017holstep}. The hyperparameters and network configurations for our approach can be found in the appendix (which is located in the supplementary material).


\paragraph{Mizar Dataset: } \emph{Mizar} \cite{mizar40for40} is a  corpus of 57,917 theorems. 
Like \cite{irving2016deepmath,olvsak2019property,kucik2018premise}, we use only the subset of 32,524 theorems which have an associated ATP proof, as those have been paired with both positive and negative premises (i.e., axioms that do / do not entail a particular theorem) to train our approach. We randomly split the 32,524 theorems as 80\% / 10\% / 10\% for training, development, and testing (yielding 417,763 / 51,877 / 52,880 individual premises). Following \cite{olvsak2019property}, each example given to the network consisted of a conjecture paired with the complete set of both positive and negative premises. The task was then to classify each individual premise as positive or negative.

%

%
\paragraph{Holstep Dataset: } \emph{Holstep} \cite{kaliszyk2017holstep} is a large corpus designed to test machine learning approaches on automated reasoning. Following prior work \cite{kaliszyk2017holstep,wang2017premise}, we use only the portion needed for proof step classification. That part has 9,999 conjectures for training and 1,411 conjectures for testing, where each conjecture is paired with an equal number of positive and negative proof steps (i.e., proof steps that were / were not part of the final proof for the associated conjecture). Using that data, we obtain 2,013,046 training examples and 196,030 testing examples, where each example is a triple with the proof step, conjecture, and a positive or negative label. We held out 10\% of the training set to be used as a development set. We follow the binary classification problem setting of \cite{wang2017premise} and \cite{kaliszyk2017holstep} where our approach classifies each proof step as either relevant or irrelevant to its paired conjecture.


\subsection{Classification Experiments}
\label{sec:classification}

\paragraph{Baselines:} For premise selection on Mizar, we compare with two existing systems: the distributed formula representation of~\cite{kucik2018premise} and the property-invariant formula representation of~\cite{olvsak2019property}. For the proof step classification task on Holstep, we compare against 4 systems implemented in two prior works: 1) DeepWalk~\cite{perozzi2014deepwalk} and FormulaNet~\cite{wang2017premise}, both of which were applied to Holstep in \cite{wang2017premise}. 2) CNN-LSTM and CNN, both introduced in the original Holstep paper \cite{kaliszyk2017holstep}.

\begin{table}[t]
\centering
\caption{Experimental results for Mizar and Holstep test sets, best result for both datasets in \textbf{bold}}
\vskip 0.1in
\label{res:exp_all}
\footnotesize
\setlength\tabcolsep{4pt}%
\begin{tabular}{ l !{\vrule} c c }
\toprule 
\textbf{Formula Embedding Method} & \textbf{Mizar Acc.} & \textbf{Holstep Acc.}  \\
\midrule
Kucik \& Korovin (2018)~\cite{kucik2018premise} & 76.5\%  & \multicolumn{1}{c}{--} \\
DeepWalk (2014)~\cite{perozzi2014deepwalk} & \multicolumn{1}{c}{--} & 61.8\%   \\
CNN-LSTM (2017)~\cite{kaliszyk2017holstep} & \multicolumn{1}{c}{--} & 83.0\%   \\
CNN (2017)~\cite{kaliszyk2017holstep}      & \multicolumn{1}{c}{--} & 82.0\%  \\
FormulaNet (2017)~\cite{wang2017premise} & \multicolumn{1}{c}{--} &  90.3\% \\
BidirDagLSTM-AttDagLSTM (this work) & \textbf{81.0}\% & \textbf{91.4}\%\\
\bottomrule
\end{tabular}
\vspace{-0.1in}
\end{table}

\paragraph{Main Results:} 
Table~\ref{res:exp_all} shows the performance for the version of our approach that incorporates the entire context surrounding a node into its embedding and jointly embeds premises / proof steps with the conjecture, i.e., a Bidirectional DAG LSTM with attention-enhanced DAG LSTM pooling.
Overall, our system outperforms all state-of-the-art systems on both datasets using a standard evaluation on held-out test data. It outperforms by a large margin on Mizar (+4.5\%, which is statistically significant with $p<0.01$) and by a moderate, but still statistically significant, margin on Holstep (+1.1\% with $p<0.01$).
In addition to the standard evaluation using a held-out test dataset reported on Table~\ref{res:exp_all}, we also compare to \cite{olvsak2019property}, which introduced a GNN designed to process specifically first-order logic theories in conjunctive normal form. In their evaluation on Mizar, they split their data into only a train and test set and evaluated the model obtained at each epoch on their test set, reporting an accuracy of ``around 80\%'' as the best performance across all test set evaluations. Following their setup, our best validation performance is 81.9\%, a roughly 2\% gain over \cite{olvsak2019property}.

The result in Table~\ref{res:exp_all} confirms our hypothesis that a more holistic treatment of logical formulae can result in a more effective embedding process than simpler methods that, by their implementation, consider less structure and embed premises and conjectures independently. 


\paragraph{Ablation Studies:} We present the results of our ablations in Table \ref{res:exp_abl}, with statistically significant ($p < 0.01$) improvements over \cite{kucik2018premise} on Mizar and \cite{wang2017premise} on Holstep marked explicitly. On Mizar, we can see that variants with attention-based pooling were the most performant by a large margin. When controlling for pooling type, the $\mathcal{R}^+$ node embedders provided better performance than the $\mathcal{R}^k$ node embedders. Similarly, when controlling for node embedding type, $\mathcal{R}^+$ pooling methods provided improvement over max pooling.

For Holstep, when controlling for node embedding type the $\mathcal{R}^+$ pooling methods had better performance than max pooling. Interestingly, when controlling for pooling type, the difference between the MPNN and $\mathcal{R}^+$ node embedding methods was not significant. Within approaches introduced here, those variants using AttDagPool did not significantly improve over those using DagPool.
We suspect that this is due to the fundamental difference between proof step classification and premise selection. Intermediate proof steps are typically much larger and noisier than actual premises, which may have led to Holstep example pairs being independent (i.e., there were properties of an individual proof step without the conjecture that would give away the positive or negative label). This is partially supported by both \cite{wang2017premise} and \cite{kaliszyk2017holstep}, who observed that their architectures performed just as well when classifying with only the proof step, rather than on both the proof step and conjecture (90.0\% vs. 90.3\% for FormulaNet and 83.0\% vs. 83.0\% for CNN-LSTM). On validation data, we also explored higher numbers of update rounds (i.e., the $k$ parameter) for variants of our approach using an MPNN as the initial node embedder; however, like \cite{wang2017premise} we found insignificant change beyond $k = 2$.

\begin{table}[t]
\centering
\caption{Ablation study on Mizar and Holstep test sets; all variations implemented by this work. Statistically significant improvements over prior work marked in \textbf{Statistically Sig.}
}
\vskip 0.1in
\label{res:exp_abl}
\footnotesize
\setlength\tabcolsep{4pt}%
\begin{tabular}{ l l c !{\vrule} c c !{\vrule} c c }
\toprule 
  \textbf{Node Embedding} & \textbf{Pool Type} & $\boldsymbol{k}$ & \textbf{Mizar Acc.} & \textbf{Statistically Sig.} & \textbf{Holstep Acc.} & \textbf{Statistically Sig.}  \\
\midrule
  MPNN & MaxPool & 2 & 76.9\% & & 90.5\%  &  \\ 
  MPNN & DagPool & 2 & 77.4\% & $\surd$ & 91.3\%  & $\surd$ \\ 
  MPNN & AttDagPool & 2 & 79.7\% & $\surd$ & 91.3\% & $\surd$ \\ 
  GCN & MaxPool & 2 & 74.7\% & & 89.0\% & \\ 
  GCN & DagPool & 2 & 77.3\% & $\surd$ & 90.9\%  & $\surd$ \\ 
  GCN & AttDagPool & 2 & 79.8\% & $\surd$ & 90.8\% & $\surd$ \\ 
  DagLSTM & DagPool & -- & 78.4\% & $\surd$ & 91.4\% & $\surd$ \\ 
  DagLSTM & AttDagPool & -- & 80.7\% & $\surd$ & 91.5\% & $\surd$  \\ 
  BidirDagLSTM & DagPool & -- & 78.1\% & $\surd$ & 91.4\%  & $\surd$  \\ 
  BidirDagLSTM & AttDagPool & -- & 81.0\% & $\surd$ & 91.4\% & $\surd$ \\ 
\bottomrule
\end{tabular}
\vspace{-0.1in}
\end{table}

\subsection{Premise Selection for Automated Theorem Proving}
\label{sec:E}
To show that our approach could be used to improve the performance of an actual theorem prover, we ran a traditional premise selection experiment with E \cite{schulz2013system}. We first trained new models using our settings from the classification experiments, however, this time optimizing for binary classification between pairs of individual formulae. 
In addition to our Mizar training set from before, we also augmented our training data by adding randomly selected negative examples for each example from our original training set. For testing, we paired the conjecture of each of the 3,252 problems from our Mizar validation set with the complete set of statements from all chronologically preceding problems (as described in \cite{irving2016deepmath}) in the union of our training and validation sets. For each problem, our model then ranked the premises with respect to each conjecture and returned the top $k \in \{ 16, 32, 64, 128, 256, 512, 1024, 2048, \infty \}$ premises (where $\infty$ indicates including all premises).

E was run on each problem in \textit{auto-schedule} mode (which tries several expert heuristics based on the given problem) with a time limit of 10 seconds per $k$, stopping at the first $k$ where the problem was solved. To validate that our approach solves more problems than E would have by itself in the same amount of time, we also measured the performance of E when run with all premises (identical to $k = \infty$) for 90 seconds per problem. Out of 3,252 problems, E by itself was able to solve 918; however, using our approach as its premise selection mechanism, E was capable of solving 1484. In both settings, E had the same amount of time (90 seconds) per problem to find a proof, but with our approach it was able to solve 566 more problems (a 61.6\% improvement) which is statistically significant with $p < 0.01$.




\bibliographystyle{unsrt}
\bibliography{all_refs}

\section{Appendix}
\label{sec:appendix}

\subsection{Network Configurations}

For Holstep, our hyperparameters were chosen to be comparable to \cite{wang2017premise}. In our model, node embeddings were 256-dimensional vectors and edge embeddings were 64-dimensional vectors. All feed-forward networks (each $F^{(t)}_{M_A}$, each $F^{(t)}_{M_C}$, each $F^{(t)}_A$, $F_{BD}$, and $F_{CL}$) followed mostly the same configuration, except for their input dimensionalities. Each had one hidden layer with dimensionality equal to the output layer (except for $F_{CL}$ where the dimensionality was half the input dimensionality). Every hidden layer for all feed-forward networks (except for $F_{CL}$) was followed by batch normalization \cite{ioffe2015batch} and a ReLU. The final activation for $F_{CL}$ was a sigmoid; for all other feed-forward networks, the final activations were ReLUs. 
For the DAG LSTMs, the hidden states were 256-dimensional vectors. Each $U^{(e_{vw})}_i$, $U^{(e_{vw})}_o$, $U^{(e_{vw})}_c$, and $U^{(e_{vw})}_f$ were learned $256 \times 256$ matrices and each of $W_i$, $W_o$, $W_f$, $W_c$, $W_a$, and $W_g$ were learned $256 \times 256$ matrices. For Mizar, all above dimensionalities were halved to be comparable to \cite{kucik2018premise,olvsak2019property}. For the attention-enhanced DAG LSTM, the multi-headed attention mechanism used two heads, with each $W^{(q)}_{i}$, $W^{(v)}_{i}$, and $W^{(k)}_{i}$ mapping from the node state dimensionality to double the node state dimensionality.

\subsection{Training}

Our models were constructed in PyTorch \cite{paszke2017automatic} and trained with the Adam Optimizer \cite{kingma2014adam} with default settings. The loss function optimized for was binary cross-entropy. We trained each model for 5 epochs on Holstep and 30 epochs on Mizar, as validation performance did not improve with more training. 
Performance on the validation sets was evaluated after each epoch and the best performing model on validation was used for the single evaluation on the test data.

\subsection{Hardware Setup}

All experiments were run on Linux machines with 72-core Intel Xeon(R) 6140 CPUs @ 2.30 GHz and 750 GB RAM, and two Tesla P100 GPUs with 16 GB GPU memory.

\end{document}